\documentclass[letterpaper, 10 pt, conference]{ieeeconf}  

\IEEEoverridecommandlockouts                              

\usepackage{graphics} 
\usepackage{epsfig} 
\usepackage{mathptmx} 
\usepackage{times} 
\usepackage{amsmath} 
\usepackage{amssymb}  
\usepackage{algorithm}
\usepackage{algorithmic}
\usepackage{multirow}
\usepackage{booktabs}
\usepackage{hyperref}
\hypersetup{
    colorlinks=true,
    linkcolor=blue,
    filecolor=magenta,
    urlcolor=blue,
    }
\title{\LARGE \bf
DT/MARS-CycleGAN: Improved Object Detection for MARS Phenotyping Robot
}

\author{David Liu\textsuperscript{*}$^{1}$, Zhengkun Li\textsuperscript{*}$^{2}$, Zihao Wu\textsuperscript{*}$^{3}$, Changying Li$^{2,3}$ 
\thanks{\textsuperscript{*}Equal Contribution.}%
\thanks{$^{1}$Athens Academy, Athens, GA 30606, USA}%
\thanks{$^{2}$University of Florida, Gainesville, FL 32611, USA}%
\thanks{$^{3}$University of Georgia, Athens, GA 30602, USA}%
\thanks{\emph{Corresponding author: Changying Li (email: cli2@ufl.edu).}}%
}

\begin{document}

\maketitle
\thispagestyle{empty}
\pagestyle{empty}

\begin{abstract}

Robotic crop phenotyping has emerged as a key technology to assess crops' morphological and physiological traits at scale. These phenotypical measurements are essential for developing new crop varieties with the aim of increasing productivity and dealing with environmental challenges such as climate change. However, developing and deploying crop phenotyping robots face many challenges such as complex and variable crop shapes that complicate robotic object detection, dynamic and unstructured environments that baffle robotic control, and real-time computing and managing big data that challenge robotic hardware/software. This work specifically tackles the first challenge by proposing a novel Digital-Twin(DT)MARS-CycleGAN model for image augmentation to improve our Modular Agricultural Robotic System (MARS)'s crop object detection from complex and variable backgrounds. Our core idea is that in addition to the cycle consistency losses in the CycleGAN model, we designed and enforced a new DT-MARS loss in the deep learning model to penalize the inconsistency between real crop images captured by MARS and synthesized images sensed by DT MARS. Therefore, the generated synthesized crop images closely mimic real images in terms of realism, and they are employed to fine-tune object detectors such as YOLOv8. Extensive experiments demonstrated that our new DT/MARS-CycleGAN framework significantly boosts our MARS' crop object/row detector's performance, contributing to the field of robotic crop phenotyping.            

\end{abstract}

\section{Introduction}
To sustain the growing world population, crop breeders strive to develop high-yielding and stress-tolerant cultivars that are more resilient to the changing climate, pests, and diseases \cite{PhenotypingRobotReview, phenotypedata, JiangyuReview, PhenotypingBotReview}. Although advances in plant biotechnology, such as genome sequencing and genotyping technologies, have significantly accelerated the pace of crop breeding process \cite{genotyping, genomeselection}, phenotyping has been a bottleneck for crop improvement \cite{PhenotypingRobotReview,CharlieLiReview}. To address this challenge, automated robotic crop phenotyping technologies have been developed to assess and quantify phenotypic traits that are related to crop growth, yield, and adaption to environmental stresses \cite{CharlieLiReview, PhenotypingBotReview, agronomy9050258}. Important crop phenotypic traits include morphological, structural, physiological, and biochemical properties at multiple organizational scales. It is crucial to phenotype a diverse range of crop genotypes across multiple environments to enable crop breeders to identify and select genotypes with desirable phenotypes best suited for specific environments. As a result, in-field high-throughput robotic phenotyping is indispensable for the effective selection of high-yielding and stress-tolerant varieties \cite{PhenotypingRobotReview, JiangyuReview, PhenotypingBotReview}.     

However, developing field crop phenotyping robots faces multiple challenges. First, to address specific phenotyping needs, in-field crop phenotyping robots need highly specific platforms such as tower-based, gantry-based, ground mobile, low- and high-altitude aerial, and satellite-based systems. These crop phenotyping platforms also need highly specific computing devices and sensors such as edge devices, RGB color cameras, depth cameras, thermal imaging, and spectral imaging, and specific software tools such as ROS (robotic operating system) and deep learning acceleration packages \cite{JiangyuReview, platforms, hardwaresoftwarereview}, to reduce the costs and allow for nondestructive characterization of the traits of interest. Second, field-based crop phenotyping robots typically need to build maps of the surrounding environment, plan their optimal paths, follow correct trajectories, navigate between crop rows, avoid obstacles, and collect phenotypical data in a breeding field. Although GPS is commonly used for robot navigation, the real-time-kinematic GPS is costly and high-quality signal reception is not always guaranteed. Therefore, the information provided by cameras along with GPS is typically necessary to precisely guide the phenotyping robot and avoid damage to crops and potential humans. Third, once large amounts of crop trait sensing datasets are acquired, data science and AI models are then needed to make necessary phenotypic inference and prediction \cite{ISUwork}. Specific image-based phenotyping may include plant stress, plant development such as plant shoot morphology and growth, plant and plant organ counting, root system architecture, and crop postharvest quality assessment, among others.  

To deal with the above-mentioned challenges, a variety of field-based high-throughput crop phenotyping robots have been reported in the literature \cite{PhenotypingRobotReview, JiangyuReview, PhenotypingBotReview, CharlieLiReview, ISUwork, Onlinedetection, MARS} to handle various types of crops, including fruits such as grapes, apples, citrus, and tomatoes, and vegetables such as sugarcane, corn, soybean, cucumber, maize, and wheat. Robotic crop phenotyping has been particularly enhanced by deep learning models and vision sensor technologies, as recently reviewed in \cite{Sensorreview}. However, developing and deploying robotic crop phenotyping systems using computer vision and deep learning still face a few key challenges, particularly in dynamic, complex, and unstructured field environments \cite{PhenotypingRobotReview}. One extensively studied while still challenging problem in the field of robotic phenotyping domain is accurate crop detection. Accurately identifying the crop target is a prerequisite for the robot to perform multiple key downstream tasks such as crop row detection, path planning, navigation, and extraction of phenotypic traits \cite{PhenotypingRobotReview, JiangyuReview, PhenotypingBotReview}.

Many deep learning based crop object detection methods have been proposed in the literature for various purposes such as crop seeding, yield estimation, phenotyping, and robotic harvesting, as surveyed in \cite{JiangyuReview, Darwin_2021, objdetectionreview}. In terms of deep learning models, convolutional neural networks (CNNs) are the most widely employed methods in crop object detection, as extensively reviewed in \cite{JiangyuReview}. Briefly, object detection CNN models were specifically designed to identify interested targets from images, such as the RCNN \cite{RCNN} and its variants based on two stages of region proposals and classifications, and the YOLO series models based on regressions of bounding boxes \cite{YOLO}. Further, these generic RCNN/YOLO models were fine-tuned on crop images with object annotations to be transferred to the agriculture domain. For instance, the RCNN model was transferred for fruit detection by fine-turning a Faster RCNN model with 100 annotated images \cite{DeepFruit}, and a comprehensive survey of this category of approaches is available in \cite{JiangyuReview}. Similarly, the YOLO model has been adopted and customized for crop detection, such as apple and pear detection in \cite{YOLOforFruit}. With the popularity of few-shot learning, many studies have explored the fine-tuning of CNN models by using several examples for crop object detection, as surveyed in \cite{fewshotlearning}. However, a generic concern in few-shot object detection is that the fine-tuned model might be subject to overfitting and its generalization capability could be a challenge, as pointed out in \cite{fewshotlearning}.

In parallel, image data augmentation strategy has been extensively explored to boost the training of deep learning models in agriculture \cite{Agimagesynthesis}. The generative adversarial network (GAN) model \cite{goodfellow2020generative} has been particularly popular in generating synthesized image data for model training, e.g., those studies in \cite{GANmodel1, GANmodel2}. The GAN model has also been widely used in the general robotics field, e.g., for synthesizing large-scale realistic images for modeling training to deal with the reality gap problem in sim-to-real transfer \cite{kleeberger2020survey,4,71, domainadaptation}. For instances, the authors in \cite{retinaGAN} developed a novel GAN model to effectively adapt synthesized images to realistic ones by enforcing an object-detection consistency for visual grasping, and the authors in \cite{rao2020rl} developed a RL-CycleGAN model by applying the similarity of total expected future reward (Q-value) of Q-learning as an additional reinforcement learning (RL)-scene consistency loss, thus enabling effective sim-to-real transfer for reinforcement learning.    

Along this promising research direction, and inspired by the recent research in \cite{DTCycleGAN}, this paper specifically proposes a novel Digital-Twin(DT)/MARS-CycleGAN model for image augmentation to improve crop object detection from complex and variable farm background in the context of our MARS robotic phenotyping \cite{MARS, Mars-Phenobot}. Our methodological innovation is that real and virtual DT MARS robots are forced to mimic each other such that the gaps between simulated and realistic robotic crop phenotyping, e.g., crop/row detection, are minimized, thus the reality gap during zero-shot sim-to-real transfer of trained crop detection models is minimized at a fundamental level. Our novel algorithmic contribution here is that in addition to the classic cycle consistency losses in the CycleGAN model \cite{zhu2017unpaired}, we designed and integrated a novel digital-twin (DT)-MARS loss in the deep learning network to penalize the inconsistency between synthesized images generated by DT MARS and real crop images sensed by MARS. As a result, the generated synthetic crop images are significantly more realistic to real images, and thus they can be used to significantly better fine-tune the object detectors such as YOLOv8 \cite{YOLOv8f}. The improved crop object detector was then employed for crop row detection in MARS' real-world robotic phenotyping environment, which is a key step for robotic path planning and navigation. Extensive experimental results have demonstrated that our DT/MARS-CycleGAN framework significantly enhanced our MARS' crop object/row detector's performance, thus contributing to the fields of crop phenotyping robotics and precision agriculture.

\section{Related Works}

\subsection{Crop Phenotyping Robot}
There have been many innovative crop phenotyping robots reported in the literature, as surveyed in \cite{JiangyuReview, PhenotypingBotReview}. As discussed in \cite{PhenotypingBotReview}, these phenotyping robots can be categorized in different ways, e.g., autonomous vs non-autonomous systems such as tractors, manually pushed or motorized carts, gantries, and cable-driven systems, and wheeled vs tracked vs wheel-legged systems. Here, we briefly introduce a few crop phenotyping robot examples that are most relevant to our work in this paper. The Thorvald II system \cite{Thorvald} was developed based on the modular hardware and software concept, that is, the robot can be easily reconfigured to obtain the necessary physical properties to operate in multiple production contexts (e.g., tunnels, greenhouses, and open fields) and the mechanical properties can be adjusted for different track widths, power requirements, and load capacities. The modular agricultural robotic system (MARS) \cite{MARS} is an autonomous and multi-purpose phenotyping robot with five essential hardware modules (wheel module, connection module, robot controller, robot frame, and power module) and three optional hardware modules (actuation module, sensing module, and smart attachment). An important feature of MARS \cite{MARS} is that different combinations of the hardware modules can make the needed configurations for specific agricultural tasks, such as crop phenotyping. The Field Scanalyzer \cite{Scanalyzer} robot platform, developed by LemnaTec, is a high-throughput phenotyping system that includes a sensor box for phenotyping sensors and environmental sensors such as cameras and LiDAR, to capture detailed phenotypic data on crops in large field plots. The Field Scanalyzer has environmental sensors on board and thus it can record climatic data during all phenotypic measurements to enable the meaningful link between phenotype and environment data. A comprehensive review of crop phenotyping robots is referred to \cite{PhenotypingBotReview}.  

\subsection{Digital Twin for Agriculture Robotics}

Digital twin (DT) has made transformative impacts on many research and industry domains \cite{liu2021review}, including agriculture \cite{DTagriculture} and robotics \cite{liu2022digital, DTCycleGAN}. The idea of real-time virtual representation via DT technologies significantly enable and facilitate the digitization of agriculture, in which data, modeling, and what-if simulations are integrated to provide a promising framework to address constraints in decision-making support and automation for various agricultural applications, including crop phenotyping, as reviewed in \cite{DTagriculture}. In parallel, DT technologies have been recently explored in robotics \cite{liu2021review,liu2022digital}. For instances, the authors in \cite{liu2022digital} reported a DT-enabled approach for achieving the effective transfer of deep reinforcement learning (DRL) algorithms to a physical robot. In \cite{DTCycleGAN}, the authors acquired large-scale visual grasping datasets with ground truth annotations in DT environment that mimic the real world to train their DT-CycleGAN model and reported promising results in zero-shot sim-to-real transfer of grasping models. In this paper, our DT/MARS-CycleGAN model takes the advantages of using DT technologies in both agriculture and robotics by maximizing the similarity between the physical and virtual DT MARS robots so that their behaviors tend to be identical, and thus the reality gap in zero-shot sim-to-real transfer is minimized.   

\subsection{Object Detection in Precision Agriculture}
Since the early work of DeepFruit \cite{DeepFruit} for crop object detection using the faster RCNN model \cite{RCNN}, the precision agriculture field has devoted much effort to improve deep learning based crop detection. For instance, in early studies, the scheme of splitting a high-resolution agriculture image into multiple small patches was proposed to deal with the CNN's small input size issue \cite{JiangyuReview}, and features from shallow layers of RCNN model were used for regional proposals to deal with the small fruit object detection problem \cite{Smallsize}. Later, few-shot learning has been employed to fine-tune the RCNN and YOLO based models by using several examples for crop object detection \cite{fewshotlearning}. Recently, vision transformer (ViT) \cite{ViTpaper} was fine-tuned for the detection of plant infections and diseases, named GreenViT \cite{ViTPlant}, and it was reported that the GreenViT achieved remarkable accuracy and effectively reduced the occurrence of false alarms \cite{ViTPlant}. The authors in \cite{chen2022maskguided} proposed a mask-guided ViT model for fruit object detection based on few-shot learning and reported promising results for apple detection. The interesting idea in \cite{chen2022maskguided} is that a mask is applied on image patches to screen out the task-irrelevant ones and to guide the ViT to focus on task-relevant and discriminative patches (e.g., apples) during few-shot learning. Particularly, the mask-guided ViT only introduces an additional mask operation and a residual connection, which enables the inheritance of parameters from pre-trained ViT without any additional cost. The authors in \cite{chen2022maskguided} also optimally selected representative few-shot samples (e.g., apples), and included an active learning based sample selection method to further improve the generalizability of the mask-guided ViT based few-shot learning. In terms of crop varieties, there are many types of deep learning models that have been developed for detecting grapes \cite{SANTOS2020105247}, apples \cite{AppleDetector}, and citrus \cite{Citrus}, among many others, as systematically surveyed in \cite{JiangyuReview}. 

However, applying deep learning techniques to agriculture applications poses unique challenges too, particularly when it comes to data collection and variability. Agricultural settings are complex and unpredictable, with factors like plant development, weather fluctuations, and variable lighting conditions, making it difficult to amass the large, diverse datasets that deep learning models require. These uncontrollable elements complicate the task of capturing a comprehensive range of data variations, limiting the deep learning technology's applicability in variable agricultural contexts. While there has been an increase in agricultural datasets recently \cite{lu2020datasetsurvey}\cite{polvara2023dataset}, there is still a pressing need for specialized, publicly available datasets tailored specifically for agricultural applications.

\subsection{Image Augmentation in Precision Agriculture}
Image data augmentation has been a mainstream approach for improving the generalization capability of deep learning models, and this effective strategy has been extensively explored to boost the training of deep learning models in agriculture \cite{Agimagesynthesis, de2022review, tobin2017domain}. Two general approaches have been proposed for image data augmentation, that is, generative deep learning models and computational simulations. The generative adversarial network (GAN) model \cite{goodfellow2020generative} and its many variants \cite{zhu2017unpaired} have been particularly widely used in generating synthesized image data for model training, e.g., those studies in precision agriculture \cite{GANmodel1, GANmodel2, khalifa2022DL-Data-augumentation-review, lu2022GAN-agricultural-review}. Early work by Giuffrida et al \cite{valerio2017DCGAN} utilized DCGAN for plant image synthesis, while Zhu et al \cite{zhu2018gan-leafcounting} leveraged Pix2Pix for improved leaf counting. Madsen et al's models \cite{madsen2019wacgan-seedling, madsen2019semi-GAN-seedlings} catered to multiple species, thus achieving a notable 64.3\% classification accuracy. Zhu et al \cite{zhu2020_cDCGAN-plantvigor} implemented cDCGAN for plant vigor rating, achieving a 23\% increase in testing accuracy, and Drees et al \cite{drees2021CGAN-brassicagrowth} used CGAN for Brassica growth predictions, revealing challenges in field applications. To address dataset scarcity, Shete et al \cite{shete2020tasselgan} introduced TasselGAN for maize images, though its realism needs enhancement. The Global Wheat-Head Dataset (GWHD) \cite{david2020_wheathead} marked an advancement for wheat phenotyping. However, domain shift challenges persisted, prompting solutions like CycleGAN by Hartley and French \cite{hartley2021cycleGAN-wheathead} which, when combined with real data, improved detection accuracy, underscoring the importance of real labeled images in agriculture. 

Computational simulations have also been widely employed in agriculture robotics to generate training data. In an early study, Rahnemoonfar and Sheppard \cite{rahnemoonfar2017tomatocount} utilized a simplistic approach for tomato counting, that is, overlaying red dots on a background. A more intricate method involves the "Cut, Paste, and Learn" strategy \cite{dwibedi2017cut}, by merging portions of distinct RGB images \cite{picon2022weedsandcorn, wang2022flowerdetection}. However, this approach is limited in representing different 3D orientations and lighting nuances. For enhanced realism, 3D plant models in virtual environments \cite{qiu2016gameengine} and techniques like photogrammetry \cite{andujar2018photogrammetry, vierbergen2023sim2realflower} and L-systems\cite{cieslak2022L-system} have been employed. Alternatively, the authors in \cite{Iqbal_2020} introduced a robotic crop phenotyping simulation system that simultaneously navigates through occluded crop rows and performs phenotyping tasks, including measuring plant volume and canopy height using a 2D LiDAR in a nodding configuration. 

However, neither GAN-based generation of synthesized images or simulation-generated images are realistic enough, which is typically called reality gap, for practical precision agriculture applications. Therefore, GAN model has been adopted for synthesizing realistic images to deal with the reality gap problem in sim-to-real transfer in the general robotics domain \cite{kleeberger2020survey,4,71, domainadaptation}. For instances, a novel RetinaGAN model \cite{retinaGAN} was developed to effectively adapt synthesized images to realistic ones by adding an object-detection consistency for visual grasping, and a RL-CycleGAN model \cite{rao2020rl} was developed by enforcing the similarity of total expected future reward (Q-value) of Q-learning as an additional reinforcement learning (RL)-scene consistency loss for visual grasping. Recently, a novel DT-CycleGAN \cite{DTCycleGAN} was proposed to further bridge the reality gap in zero-shot sim-to-real transfer of visual grasping models. In DT-CycleGAN, in addition to the traditional cycle consistency losses, a grasping agent's action consistency loss was defined and penalized to minimize the inconsistency of the grasping agent's actions between the virtual states generated by the DT-CycleGAN generator and the real visual states. A key innovation of DT-CycleGAN is that it effectively integrates two mainstream image data augmentation methods of generative CycleGAN model and simulations in the digital twin space. This DT-CycleGAN model inspired us to perform simulations to generate synthesized crop images in the virtual space and employ a novel DT-MARS consistency loss to make the synthesized crop images more similar to real-world ones sensed by our MARS phenotyping robot.

\begin{figure*}[!ht]
\centering
\includegraphics[width=0.89\textwidth]{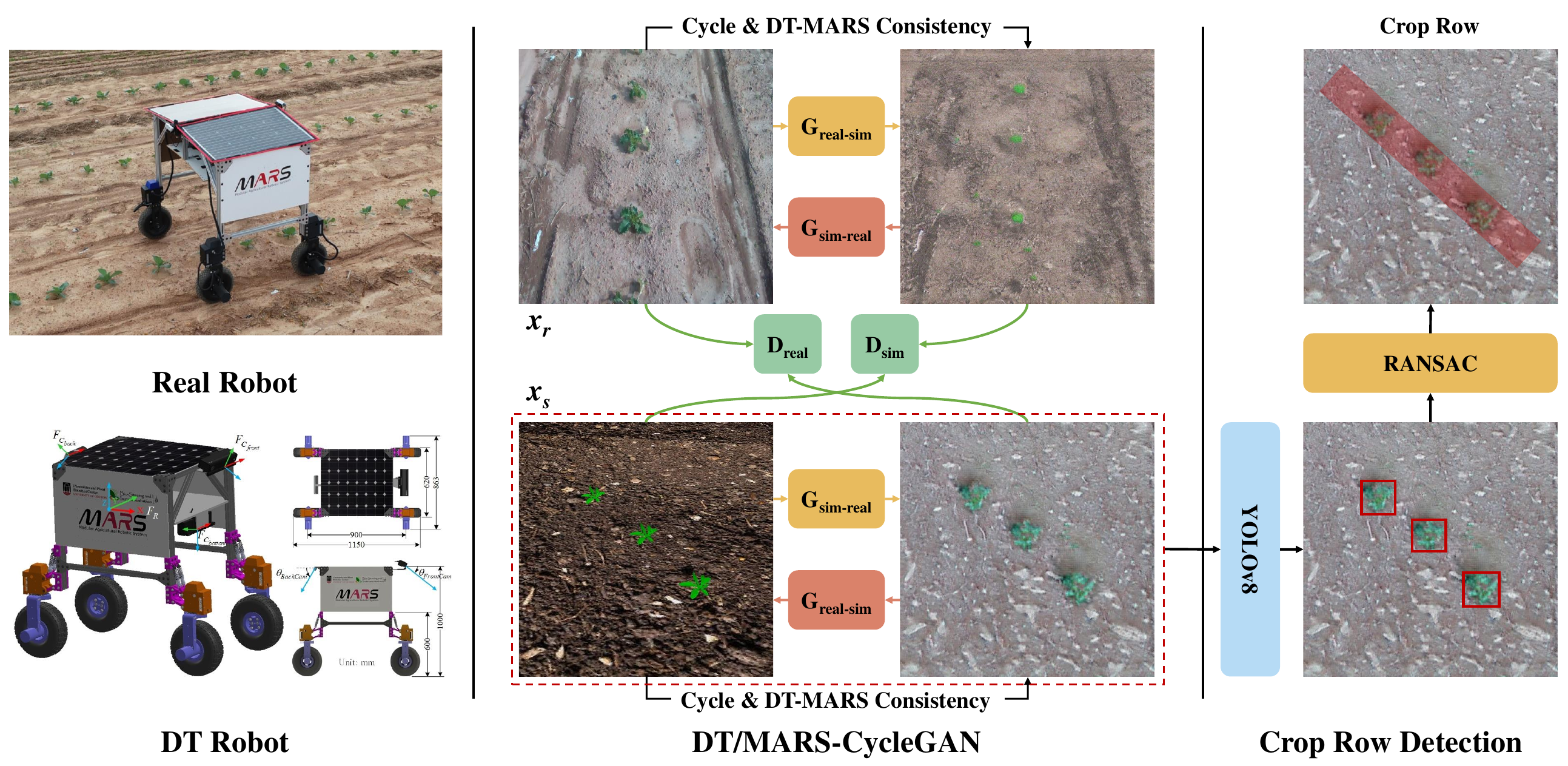}
\caption{Illustration of the physical/DT robots (left panel), the DT/MARS-CycleGAN model (middle panel), and the crop object/row detection network (right panel). Additional annotation and description are referred to the main text.} 
\label{main}
\end{figure*}

\section{Methods}

As illustrated in Figure \ref{main}, our framework is comprised of three main components: the robots operational in both real-world and simulated environments, the crop object/row detection network; and the DT/MARS-CycleGAN model that integrates the robot and the object detection network.

\subsection{Physical and Digital-Twin (DT) Robots}
\label{sec:Physical/DT robots}

Figure \ref{main} (left panel) showcases the MARS phenotyping robot (PhenoBot) \cite{Mars-Phenobot} that is employed in this work. The MARS PhenoBot is a solar-powered modular platform with a four-wheel steering and four-wheel driving configuration. Each wheel module of MARS PhenoBot is equipped with an independent suspension mechanism, which makes it adaptable to uneven field terrain. The MARS PhenoBot robot is designed for streamlined modular phenotyping and specifically tailored for crop phenotyping \cite{MARS, Mars-Phenobot}. The robot is outfitted with three cameras, capturing views from the front, rear, and bottom, respectively. This setup allows for effective monitoring of crop growth and offers crucial row-specific data to enhance farm field navigation and to prevent crop damage. The bottom camera is specifically pointed at the ground to observe more details of the farm field and to detect crops and rows. More details of the MARS PhenoBot are referred to \cite{Mars-Phenobot}. The virtual DT robot of MARS PhenoBot was developed using the Solidworks tool. Special attention was given to simulating the camera perspectives (e.g., bottom camera for crop/row detection) to reduce discrepancies in visual sensing between the real robot and the DT version, thereby narrowing the reality gap during zero-shot sim-to-real transfer of crop object detection models.

\subsection{Crop Object/Row Detection Network}
The crop object/row detection network includes two components: initially, crop detection is facilitated by the YOLOv8 object detection model \cite{YOLOv8f}, which is then followed by the application of the RANSAC (RANdom SAmple Consensus) algorithm \cite{RANSAC} for delineating crop rows.

As depicted in Figure \ref{main}, crop field images coupled with corresponding crop object annotations are initially fed into the YOLOv8 model. The input image is represented as $ \mathbf x\in  \mathbb{R} ^{B2CHW}$, where $BCHW$ denotes the batch size (B), channel (C), height (H), and width (W) of the images. For expedited training, images are resized to 
224×224. The bounding box annotation is formatted with one row per object represented as $ \mathbf y\in  \mathbb{R} ^{Bn4}$, where $B$ is the batch size, $n$ represents the number of objects in an image, each described in $[x_{center}, y_{center}, width, height]$ format. By optimizing the IoU loss between the predicted bounding boxes and the ground truth, the YOLOv8 model can accurately locate the crop objects in the images. 

With the crop bounding boxes predicted by YOLOv8, the centers of all bounding boxes are extracted together to form the points data essential for crop row detection. Subsequently, the RANSAC algorithm is employed to discern a line representing the crop row that best fits the center points of the bounding boxes of crop objects. Then the detected crop row will be used as the input for visual serving control of the MARS PhenoBot. That is, the error offset of the detected crop row from the robot's reference is used as the signal input to a PID (proportion, integration, and differentiation) controller to control the robot’s velocity and movement \cite{Mars-Phenobot}. It is apparent that accurate detection of crops and crop rows is a fundamentally important task for crop phenotyping robot's path planning and navigation.   

For zero-shot sim-to-real transfer of the crop detection model trained in a simulated environment, the object detection model along with both the images and the bounding box predictions are holistically modeled by the following DT/MARS-CycleGAN model.

\subsection{DT/MARS-CycleGAN: Cycle Consistency Loss}
The CycleGAN introduced by Zhu et al. \cite{zhu2017unpaired} was originally developed for unpaired image style transfer. In our work, it is adapted and integrated into the DT/MARS-CycleGAN model to address the discrepancies in visual appearance between simulated and real-world images, as demonstrated by the real-to-sim and sim-to-real transitions in the middle panel of Figure \ref{main}. Here, $\mathbf{x_r} \in \mathbf{R}$ represents real visual images, and $\mathbf{x_s} \in \mathbf{S}$ denotes simulated visual images, as illustrated in Figure \ref{main}. The simulated images are acquired from the simulation environment, created by utilizing farm field images collected from the Internet as background and rendering 3D models of the crops as target objects at random positions. A virtual camera mounted on the DT robot captures these images within the simulated environment, and an analytical virtual model extracts the ground truth bounding boxes for each crop in the image based on predefined position information. Meanwhile, real images were obtained from actual farm environments by capturing video streams as the real robot traversed along the crop rows in the farm field.

Specifically, as delineated in the DT/MARS-CycleGAN panel in Figure \ref{main}’s center panel, the generator, $G_r(\cdot)$, transforms images to a real style, and $G_s(\cdot)$ modifies the input images to a simulation style. These are correspondent to the generators $F(*)$ and $G(*)$ in Zhu et al’s original CycleGAN model \cite{zhu2017unpaired}. The discriminator, $D_r(\cdot)$, evaluates whether the input images are in real style, while $D_s(\cdot)$ discerns if the images exhibit a simulation style. The cycle loss function of DT/MARS-CycleGAN is mathematically described as:
\begin{align}
L(G_s,G_r,D_s,D_r) = & L_{GAN}(G_s,D_s,\mathbf{R}, \mathbf{S})+L_{GAN}(G_r,D_r, \mathbf{S}, \mathbf{R})\notag\\
+& L_{cyc}(G_r,G_s) + L_{identity}(G_r,G_s)
\end{align}

Here, $L_{GAN}(G_s,D_s,\mathbf{R},\mathbf{S})$ and $L_{GAN}(G_r,D_r,\mathbf{S}, \mathbf{R})$ denote the adversarial losses; $L_{cyc}(G_r,G_s)$ signifies the cycle consistency loss, and $L_{identity}(G_r,G_s)$ represents the identity mapping loss.

Adversarial losses aim to align the style of the generated crop image with the style of the target domain, such as in a real-world farm environment. They are expressed as:
\begin{align}
L_{GAN}(G_s,D_s,\mathbf{R},\mathbf{S}) = &\mathbb{E}{\mathbf{x_s} \sim p{data}(\mathbf{S})}[log D_s(\mathbf{x_s})] \\
\notag +&\mathbb{E}{\mathbf{x_r} \sim p{data}(\mathbf{R})}[log (1 - D_s(G_s(\mathbf{x_r})))] \\
L_{GAN}(G_r,D_r,\mathbf{S},\mathbf{R}) = &\mathbb{E}{\mathbf{x_r} \sim p{data}(\mathbf{R})}[log D_r(\mathbf{x_r})] \\
\notag +&\mathbb{E}{\mathbf{x_s} \sim p{data}(\mathbf{S})}[log (1 - D_r(G_r(\mathbf{x_s})))]
\end{align}

To inhibit the mapping of the input image to any arbitrary permutation of images in the target domain style, the cycle consistency loss is utilized, ensuring that the images can revert to their original form within, for instance, a virtual environment. This is defined as:
\begin{align}
L_{cyc}(G_r,G_s) = &\mathbb{E}{\mathbf{x_s} \sim p{data}(\mathbf{S})}[\Vert G_s(G_r(\mathbf{x_s}))-\mathbf{x_s} \Vert_1] \\
\notag + &\mathbb{E}{\mathbf{x_r} \sim p{data}(\mathbf{R})}[\Vert G_r(G_s(\mathbf{x_r}))-\mathbf{x_r} \Vert_1]
\end{align}

Lastly, the identity mapping loss preserves content consistency within the crop image and is formulated as:
\begin{align}
L_{identity}(G_r,G_s) = &\mathbb{E}{\mathbf{x_s} \sim p{data}(\mathbf{S})}[\Vert G_s(\mathbf{x_s})-\mathbf{x_s} \Vert_1] \\
\notag + &\mathbb{E}{\mathbf{x_r} \sim p{data}(\mathbf{R})}[\Vert G_r(\mathbf{x_r})-\mathbf{x_r} \Vert_1]
\end{align}

\subsection{DT/MARS-CycleGAN: DT-MARS Consistency loss}
Inspired by the DT-CycleGAN model \cite{DTCycleGAN} that was originally designed for visual grasping, a novel DT-MARS consistency loss is introduced here to facilitate and enforce more coherent sim-to-real transfer of crop images. This additional loss aims to penalize any disparities in the positions and sizes of the DT crops represented by the bounding boxes on the image between the original and the style-transformed image, which is the output of the generator.

Given an image, denoted as $\boldsymbol{x} \in \boldsymbol{S}$, the position and size of the crops, represented as $\boldsymbol{a}_1 = \text{Detector}(\boldsymbol{x})$, should ideally correspond with $\boldsymbol{a}_2 = \text{Detector}(G_r(\boldsymbol{x}))$. Therefore, the losses associated with DT-MARS consistency, which quantitatively assess the discrepancies in the predicted bounding boxes as depicted in the middle panel of Figure \ref{main}, are formulated as:
\begin{align}
&L_{DT-MARS}(G_r, G_s, \text{Detector}) \notag\\
&=\mathbb{E}{\boldsymbol{x_s} \sim p{\text{data}}(\boldsymbol{S})}[\Vert \text{Detector}(G_r(\boldsymbol{x_s})) - \text{Detector}(\boldsymbol{x_s}) \Vert_1] \notag\\
&+\mathbb{E}{\boldsymbol{x_r} \sim p{\text{data}}(\boldsymbol{R})}[\Vert \text{Detector}(G_s(\boldsymbol{x_r})) - \text{Detector}(\boldsymbol{x_r}) \Vert_1]
\end{align}

Thus, the total loss function for DT/MARS-CycleGAN model is represented as:
\begin{align}
\label{eq_dtcyclegan}
&L(G_s, G_r, D_s, D_r, \text{Detector}) \notag\\
&= \lambda_{\text{gan}} * L_{\text{GAN}}(G_s, D_s, \boldsymbol{R}, \boldsymbol{S}) + \lambda_{\text{gan}} * L_{\text{GAN}}(G_r, D_r, \boldsymbol{S}, \boldsymbol{R}) \notag\\
&+ \lambda_{\text{cyc}} * L_{\text{cyc}}(G_r, G_s) + \lambda_{\text{identity}} * L_{\text{identity}}(G_r, G_s) \notag\\
&+ \lambda_{\text{detector}} * L_{\text{DT-MARS}}(G_r, G_s, \text{Detector})
\end{align}

In this expression, $\lambda_{\text{gan}}$, $\lambda_{\text{cyc}}$, $\lambda_{\text{identity}}$, and $\lambda_{\text{detector}}$ are the hyperparameters governing the impact of the respective loss components in the overall loss function. In our model, the values experimentally assigned to these hyperparameters are $\lambda_{\text{gan}} = 1$, $\lambda_{\text{cyc}} = 5$, $\lambda_{\text{identity}} = 2$, and $\lambda_{\text{detector}} = 10$.

The introduction of the DT-MARS consistency loss ensures that the bounding boxes, representing the DT crops in both of the real and simulated domains, are coherent in terms of their positions and dimensions, leading to more robust and accurate sim-to-real transfer by mitigating the discrepancies between these two domains.

\subsection{Training Strategies}
The training process of the complete framework is divided into two stages: the first step involves the training of the crop detection model, and the second engages the DT/MARS-CycleGAN for zero-shot sim-to-real transfer. Both training stages are conducted exclusively within simulated environments, without the need for human annotation on the real world farm images.

\textbf{Training Crop Detection Model:}
Initially, we constructed the simulation environment by utilizing the Pybullet package, employing real farm field images as the background for the simulation environment, and arbitrarily placing 3D models of crops as target objects at random positions. To enrich the diversity of the generated simulation images, we collected 500 distinct publicly available images of farm fields and created 3D models of three different crop species: sweet beet, polygonum and cirsium at varying growth stages. Subsequently, images are captured by utilizing a virtual camera positioned on the DT robot (bottom view for crop/row sensing) within the simulation environment, and an analytical virtual model extracts the ground truth bounding boxes for each crop in the image based on the predefined position information.

The generated simulation crop detection dataset with mimicked appearances and accurate bounding box annotations is then employed for crop detection model training. Given the hardware constraints of real crop robot and the necessity for real-time processing in robotic phenotyping tasks, we employ the YOLOv8n detection model for lightweight deployment and swift inference. Specifically, the YOLOv8n weights pre-trained on the COCO dataset serve as the initialization, followed by a fine-tuning step on our generated simulation crop detection dataset for domain adaptation.

\textbf{Training DT/MARS-CycleGAN:}
By incorporating unlabeled real image frames extracted from video streams captured by the real robot that navigated along crop rows in actual farm environments, 
the reality gap in zero-shot sim-to-real transfer is notably reduced during the training of DT/MARS-CycleGAN within the simulation environment. Specifically, 2,400 simulation images with automatically annotated bounding boxes, and 2,400 real frames extracted from the video stream are employed. The training configurations align with those described in \cite{zhu2017unpaired}. In contrast to RetinaGAN \cite{retinaGAN}, where the sim-to-real transfer process solely functions as a method to augment data for training the detection model, our framework facilitates the simultaneous training of the crop detection model and the DT/MARS-CycleGAN. Subsequent experiments reveal that this novel methodology achieves superior results when compared to the RetinaGAN model \cite{retinaGAN}.

\section{Evaluation}
\subsection{Dataset}
The training set is comprised of images from two domains: 2,400 auto-annotated simulation crop detection data generated from the PyBullet simulation environment, and 2,400 real frames extracted from the video stream as previously detailed. The simulation data comprises three crop species: sweet beet, polygonum, and cirsium at different growth stages, while the real data includes two crop species: collard and kale. To assess the performance of the trained model, we assembled a high-quality test set including 408 unseen image frames gathered by the real robot in farm fields. Each image is manually labeled by utilizing the RainbowFlow labeling tool, without data augmentation, and is formatted in the YOLO object detection dataset format.

\subsection{Zero-shot Sim-to-Real Transfer}
We assess four distinct zero-shot sim-to-real transfer methods on the created dataset:

\begin{itemize}
\item Sim-Only: Training solely on simulation data, without any sim-to-real transfer.
\item CycleGAN: Training on sim-to-real transferred images generated by CycleGAN.
\item RetinaGAN: Training on sim-to-real transferred images with additional perception consistency, generated by RetinaGAN.
\item The proposed DT/MARS-CycleGAN.
\end{itemize}

\begin{figure}[!ht]
\centering
\includegraphics[width=0.48\textwidth]{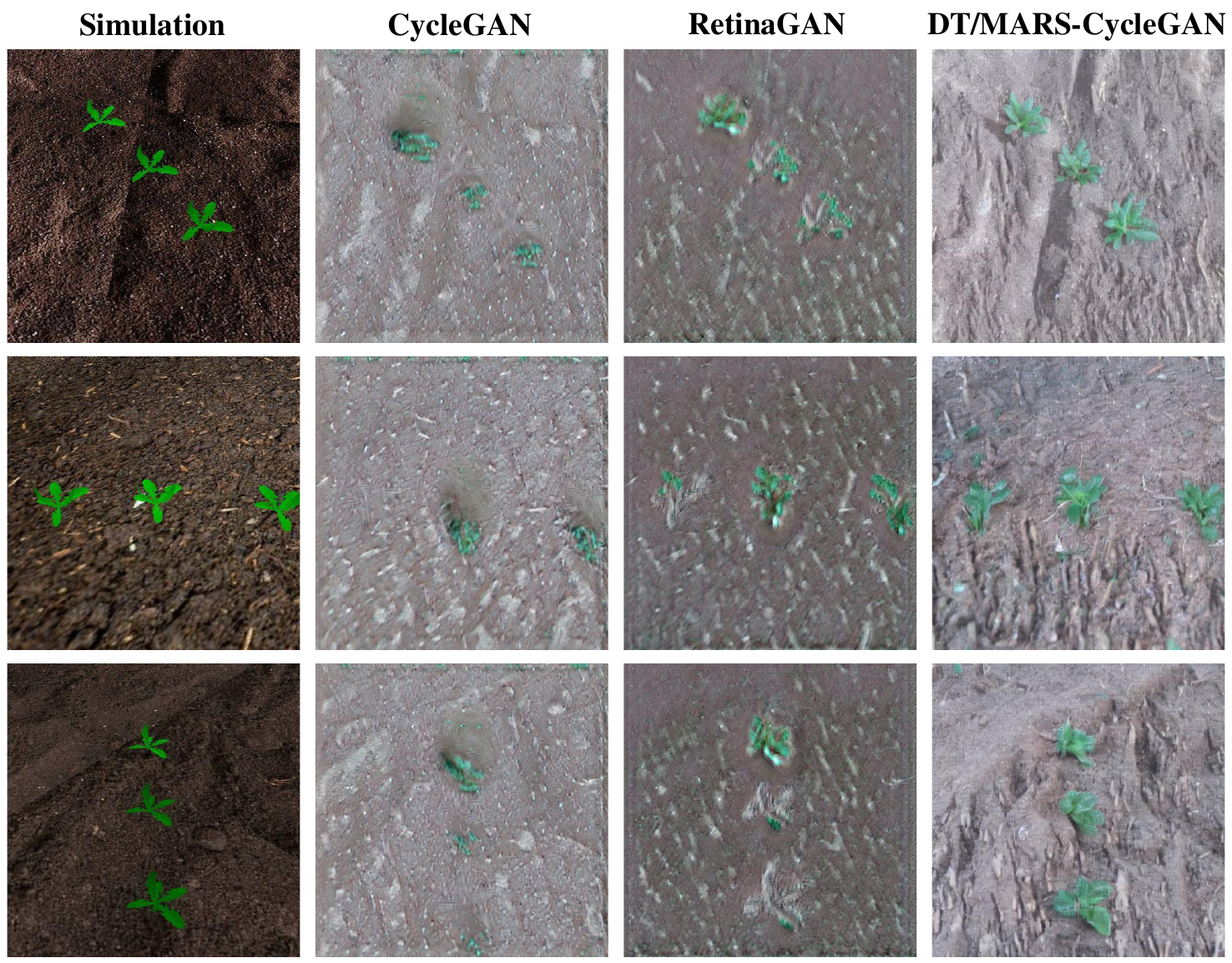}
\caption{Simulation-to-Real synthetic images generated by different models.} 
\label{synthetic}
\end{figure}

The crop detection model we utilize is YOLOv8n, which is chosen for lightweight deployment and the necessity of real-time inference in real-world agricultural applications. In all four sim-to-real transfer methods, the YOLOv8n detection model utilizes the pretrained weights on the COCO dataset for initialization and are trained for 100 epochs with a batch size of 64, learning rate of 1e-2, weight decay of 5e-4, a cosine learning rate scheduler, and the IoU loss function. All four models are trained exclusively in simulation environments and are never fine-tuned on labeled real-world data to ensure zero-shot sim-to-real transfer.

Figure \ref{synthetic} displays the visualizations of generated sim-to-real images by different methods, indicating that although CycleGAN may effectively transfer background information within an image, information pertinent to the crop target is often distorted or even lost. Conversely, with the additional consistency loss of target detection, RetinaGAN can somewhat adequately preserve target information; however, the background transformed by this model appears uniform and repetitive, lacking in variety. This could be due to the fixed detection model parameters in RetinaGAN, which constrain the model within the simulation domain and hamper its adaptability in the real domain, leading to an incomplete transfer of background style. In contrast, DT/MARS-CycleGAN showcases superior performance by enabling the concurrent training of the crop detection model and the DT/MARS-CycleGAN model, thereby emphasizing the effectiveness of our framework in addressing reality gaps in zero-shot sim-to-real transfer scenarios.

\begin{figure}[!ht]
\centering
\includegraphics[width=0.48\textwidth]{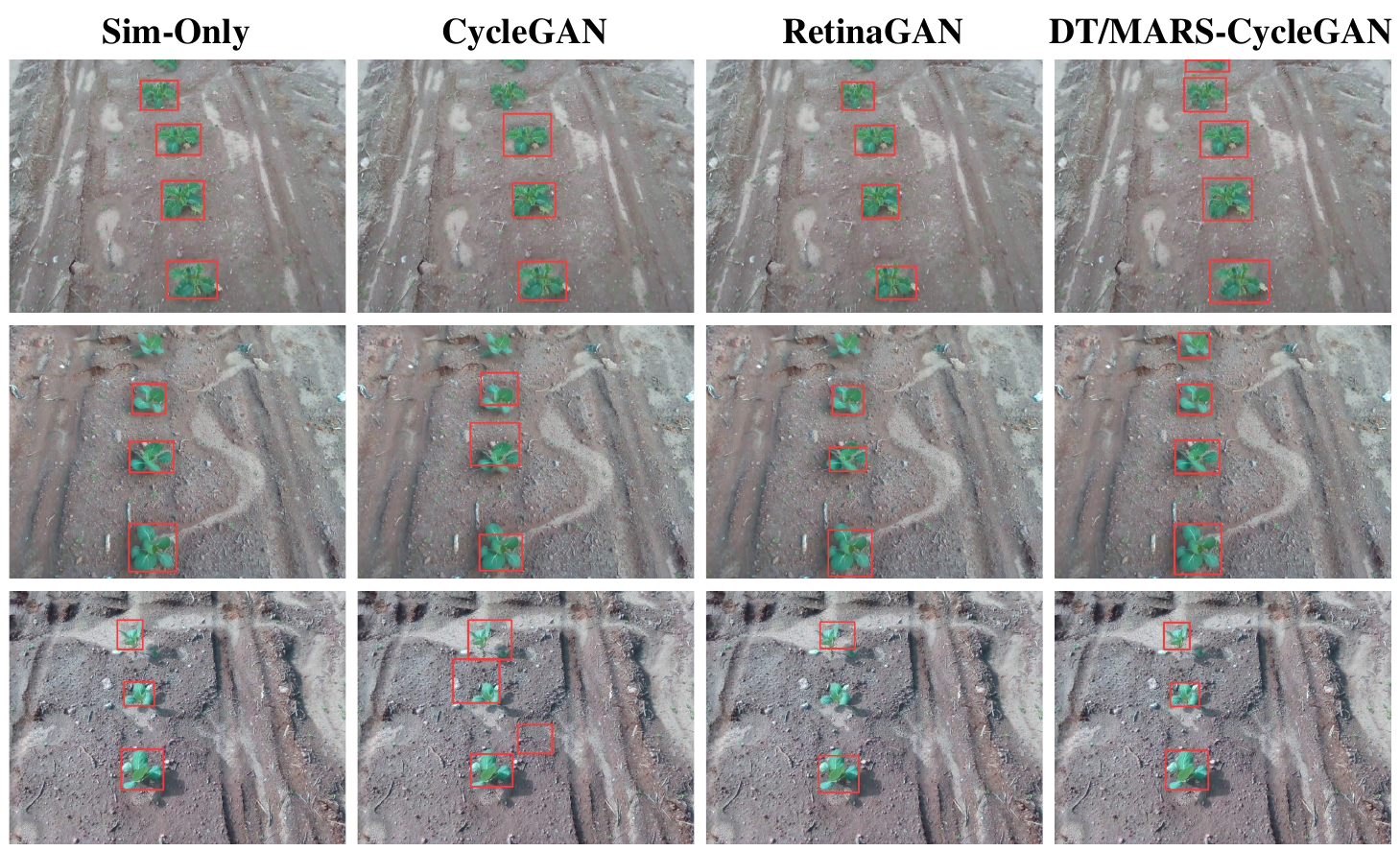}
\caption{Visualization of crop detection across different methods.} 
\label{detection view}
\end{figure}

Therefore, as shown in Table \ref{sim2real result}, the proposed DT/MARS-CycleGAN achieves superior performance on the test dataset collected in the real-world environment, demonstrating promising zero-shot sim-to-real transfer capability. In contrast, due to the target position shift and monotonous background appearance, both the CycleGAN and RetinaGAN methods yield even poorer performance than Sim-Only. Figure \ref{detection view} provides qualitative visualizations of crop detection for these methods. The proposed DT/MARS-CycleGAN model effectively reduces both false positives and false negatives in comparison to other methods.

\begin{table}[ht]
\caption{The crop detection results of YOLOv8 with different sim-real transfer methods.}
\label{sim2real result}
\begin{center}
\begin{tabular}{lcccc}
\toprule
Methods & P & R & mAP50 & mAP50-95 \\
\midrule
Sim-Only         &  0.925 & 0.827 & 0.917 & 0.656 \\
CycleGAN          & 0.500 & 0.422 & 0.411 & 0.156 \\
RetinaGAN          & 0.741 & 0.573 & 0.650 & 0.221 \\
DT/MARS-CycleGAN      & 0.942 & 0.895 & \textbf{0.964} & \textbf{0.674} \\       
\bottomrule
\end{tabular}
\end{center}
\end{table}

\begin{table}[ht]
\caption{The crop row detection results with different sim-to-real transfer methods.}
\label{row}
\begin{center}
\begin{tabular}{ccccc}
\toprule
\multirow{2}{*}{Method} & \multicolumn{2}{c}{MAE with RANSAC}&\multicolumn{2}{c}{MAE with Linefit}   \\ \cline{2-5} 
                        & Angle(deg)   & Dist(px) & Angle(deg)    & Dist(px) \\ 
\midrule
Sim-Only                & 2.47          & 4.34             & 1.23          & 2.55          \\
CycleGAN                & 9.4           & 32.18            & 8.83          & 27.84         \\
RetinaGAN                & 8.05           & 22.56           & 7.20          & 27.49         \\
DT/MARS         & \textbf{1.67} & \textbf{3.33}    & \textbf{1.00} & \textbf{2.44} \\ 
\bottomrule
\end{tabular}
\end{center}
\end{table}

\subsection{Crop Row Detection}
\begin{figure}[h]
\centering
\includegraphics[width=0.48\textwidth]{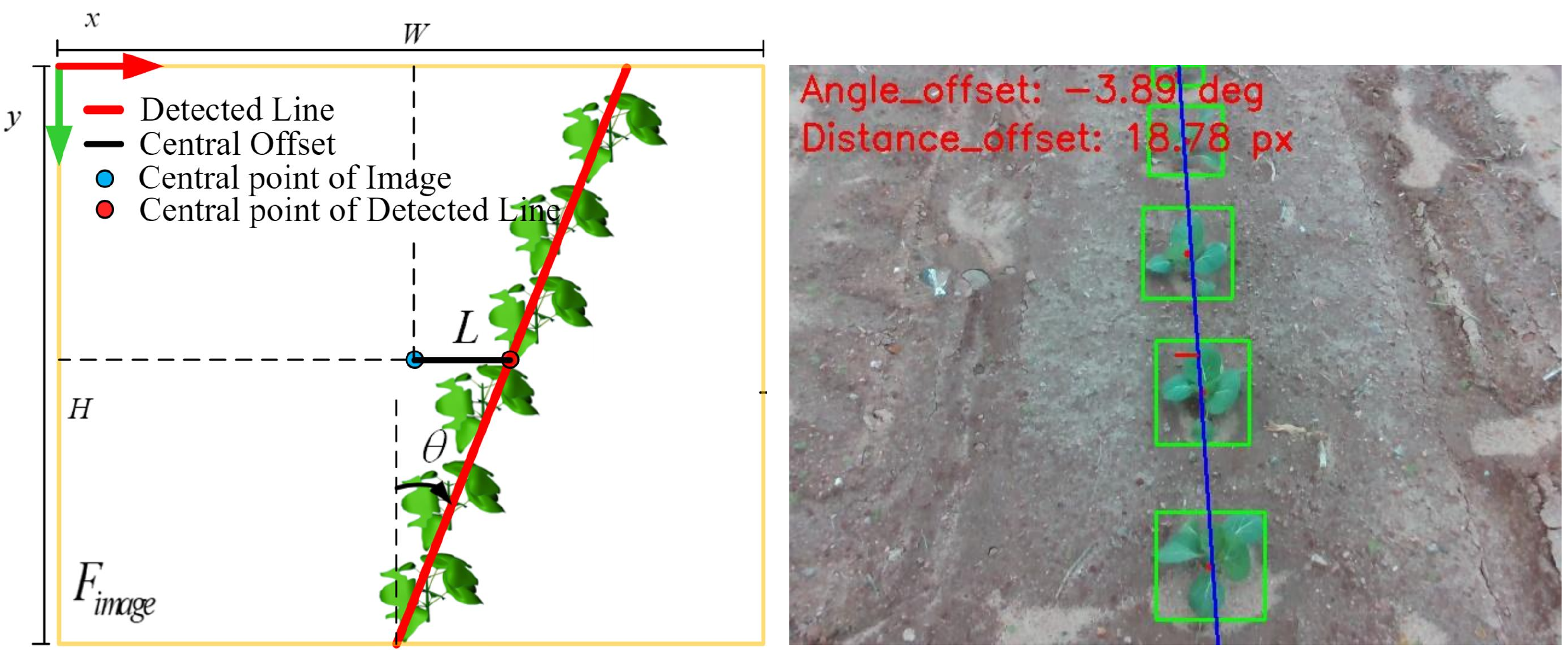}
\caption{Illustration of the crop row detection. (a) shows the two primary parameters to evaluate the row detection: angular offset and distance offset; (b) is an example of row detection in the real farm field.}
\label{row_detection_illumination}
\end{figure}

Crop row detection is crucial for determining the direction of movement in the robot's visual navigation, as well as providing supplemental visual constraints for a multi-sensor fusion navigation approach, as shown in prior studies \cite{Mars-Phenobot}. In the context of multiple visual servo controllers, the crop row detection process can be viewed as a sensor. This visual sensor consistently generates signals depicting the error offset from a reference, as illustrated in Figure \ref{row_detection_illumination}:

\textbf{Angular Offset $\theta$}: represents the angle at which the detected crop line that deviates from the vertical orientation.

\textbf{Horizontal Offset L}: denotes the discrepancy between the image's central point and the detected line's center point.

In the crop row detection evaluation, the four trained models with synthesized data (Sim-only, CycleGAN, RetinaGAN, and DT/MARS-CycleGAN) were first used for crop detection and to calculate the central points of each crop. Then two common line fitting algorithms including RANSAC and basic line fitting algorithm with the least squares method were used to calculate the two offset parameters $\theta$ and L. Finally, we calculated the offset difference between the ground truth and the predicted one with the trained models.

The mean absolute error (MAE) of offset difference was used to evaluate the crop row detection. As shown in Table \ref{row}, our proposed DT/MARS-CycleGAN exhibits less crop row detection error in both angular offset error and horizontal offset error. The main reason is that the crop detector trained with DT/MARS-CycleGAN generated images possesses better zero-shot transfer ability in real-world farm field situations. In addition, the overall performances of the basic line fitting algorithm with the least squares method are better than RANSAC, and the main reason is that the number of crops in testing images are mainly in the range of 3 to 6 and the RANSAC-based algorithm sometimes could filter abnormal values that actually belong to the real crops. The improved crop row detection method will be deployed for automated navigation of the MARS PhenoBot in real-world crop farm field in the future.

\begin{table}[ht]
\caption{The crop detection results of YOLOv8 with different numbers of training samples in simulation.}
\label{training sample}
\begin{center}
\begin{tabular}{lcccc}
\toprule
\# of training samples & P & R & mAP50 & mAP50-95 \\
\midrule
100         &  0.969 & 0.739 & 0.903 & 0.607 \\
300          & 0.933 & 0.724 & 0.899 & 0.606 \\
600          & 0.937 & 0.803 & 0.899 & 0.641 \\
1200          & 0.923 & 0.808 & 0.902 & 0.649 \\
2400      & 0.925 & 0.827 & \textbf{0.917} & \textbf{0.656} \\     
\bottomrule
\end{tabular}
\end{center}
\end{table}

\begin{table}[ht]
\caption{The crop detection results of YOLOv8 trained on simulation images of different crop species.}
\label{crop species}
\begin{center}
\begin{tabular}{lcccc}
\toprule
Crop & P & R & mAP50 & mAP50-95 \\
\midrule
Sugar beet         & 0.909 & 0.802 & 0.889 & 0.612 \\
Polygonum          & 0.846 & 0.696 & 0.807 & 0.559 \\
Cirsium          & 0.966 & 0.616 & 0.885 & 0.585 \\
Mixed          & 0.925 & 0.827 & \textbf{0.917} & \textbf{0.656} \\ 
\bottomrule
\end{tabular}
\end{center}
\end{table}

\begin{table}[ht]
\caption{The crop detection results of YOLOv8 trained on simulation images of sugar beet at different growth stages.}
\label{crop size}
\begin{center}
\begin{tabular}{lcccc}
\toprule
Crop & P & R & mAP50 & mAP50-95 \\
\midrule
Seedling         & 0.886 & 0.552 & 0.748 & 0.452 \\
Well-grown          & 0.907 & 0.802 & 0.878 & 0.599 \\
Mixed          & 0.909 & 0.802 & \textbf{0.889} & \textbf{0.612} \\ 
\bottomrule
\end{tabular}
\end{center}
\end{table}

\begin{table*}[ht]
\caption{The crop detection results of YOLOv8 trained on simulation images \\ of different backgrounds and numbers of objects per image.}
\label{background}
\begin{center}
\begin{tabular}{lcccc}
\toprule
Setting & P & R & mAP50 & mAP50-95 \\
\midrule
single object + single background         & 0.512 & 0.466 & 0.516 & 0.339 \\
single object + multi-background          & 0.238 & 0.905 & 0.849 & 0.550 \\
multi-object + single background          & 0.607 & 0.927 & 0.910 & 0.613 \\
multi-object + multi-background          & 0.925 & 0.827 & \textbf{0.917} & \textbf{0.656} \\ 
\bottomrule
\end{tabular}
\end{center}
\end{table*}

\subsection{Ablation Studies}
\textbf{Evaluation on Different Training Data Sizes:}
Our proposed DT/MARS-CycleGAN model significantly diminishes the human effort required to accumulate labeled real-world training data of annotated crop object bounding boxes. By eliminating this laborious endeavor, we can harness substantial amounts of simulation data to enhance zero-shot sim-to-real transfer further. To quantify the influence of the size of simulation data, we trained the crop detection model with varying sample sizes in the simulation environment. As depicted in Table \ref{training sample}, the mAP50 elevates from 0.903 to 0.917 by merely augmenting simulation data from 100 to 2,400, and the mAP50-90 ascends from 0.607 to 0.656. These findings underscore that ample simulation data can notably boost detection performance and consequently aid in achieving high-quality sim-to-real transfer by contributing to the DT-MARS consistency loss in the proposed DT/MARS-CycleGAN.

\textbf{Evaluation on Different Crop Species and Growth Stages:}
Crop species and growth stages can significantly influence a model's generalizability, as appearances across different species and growth stages can be either highly similar or distinctly different. Selecting an optimal set of crop species and growth stages for generating simulation images can improve the performance of crop detection models pre-trained on such simulation data. In Table \ref{crop species}, we trained the detection model on simulation images with various species. Results indicate that sugar beet achieves better detection performance compared to other species. This is likely because its appearance closely resembles that of collard and kale in the test images. Additionally, training data that combines various species outperforms data from any single species, suggesting that diverse species combinations can enrich training data and thus enhance model generalizability. Table \ref{crop size} presents the detection performance based on training with different growth stages. Similarly, mixed data covering all growth stages surpasses training that focuses solely on one stage.

\textbf{Evaluation on Different Backgrounds and Number of Objects per Image:}
The images generated in the simulation environment are pivotal to the proposed framework as they compose the annotated crop images for both stages of the training. To evaluate the effectiveness of varying generation settings in the simulation environment, we conducted object detection model training using different combinations presented in Table \ref{background}. As depicted in Table \ref{background}, the object detection model achieves enhanced performance with a substantial number of field images used as environment backgrounds and multiple 3D crop models, underscoring the high diversity and quality of the simulation images generated by our proposed framework.

\section{Conclusions and Discussions}
This paper introduces a novel DT/MARS-CycleGAN framework, effectively bridging the reality gap between simulated and real-world environments and facilitating effective zero-shot sim-to-real transfer in robotic crop detection. By imposing both cycle and DT-MARS consistency losses, which penalizes discrepancies in visual appearance and crop target between synthesized and real-world crop images, the proposed method achieves highly effective sim-to-real transfer. The fine-tuned object detectors on this diverse and high-quality synthetic data substantially elevates detection performance. Extensive experiments demonstrate that the proposed DT/MARS-CycleGAN framework enables more robust crop perception from complex backgrounds, advancing the field of robotic crop phenotyping. This work provides an effective solution to a critical robotic vision challenge in unstructured and complex agricultural environments, moving towards more productive and resilient crop phenotyping and crop breeding process in the future.

This work can be further extended along the general direction of foundation models and large vision models \cite{bommasani2021opportunities, zhou2023comprehensive} in the future. Though our DT/MARS-CycleGAN framework demonstrated a reasonably good zero-shot transfer capability, its performance in dealing with the variation of real-world crop images can be further improved using the new methodology of foundation model \cite{bommasani2021opportunities}. To realize such a goal, the backbones of our DT/MARS-CycleGAN and object detection networks would be upgraded by the more powerful ViT model and its variants \cite{ViTpaper, chen2022maskguided, kirillov2023segment}, and should be trained and fine-tuned on much larger scale of datasets, towards a foundation model of crop image understanding and other tasks in the future.








\bibliographystyle{IEEEtran}
\bibliography{ref}

\begin{thebibliography}{10}
\providecommand{\url}[1]{#1}
\csname url@samestyle\endcsname
\providecommand{\newblock}{\relax}
\providecommand{\bibinfo}[2]{#2}
\providecommand{\BIBentrySTDinterwordspacing}{\spaceskip=0pt\relax}
\providecommand{\BIBentryALTinterwordstretchfactor}{4}
\providecommand{\BIBentryALTinterwordspacing}{\spaceskip=\fontdimen2\font plus
\BIBentryALTinterwordstretchfactor\fontdimen3\font minus
  \fontdimen4\font\relax}
\providecommand{\BIBforeignlanguage}[2]{{%
\expandafter\ifx\csname l@#1\endcsname\relax
\typeout{** WARNING: IEEEtran.bst: No hyphenation pattern has been}%
\typeout{** loaded for the language `#1'. Using the pattern for}%
\typeout{** the default language instead.}%
\else
\language=\csname l@#1\endcsname
\fi
#2}}
\providecommand{\BIBdecl}{\relax}
\BIBdecl

\bibitem{PhenotypingRobotReview}
A.~Atefi, Y.~Ge, S.~Pitla, and J.~Schnable, ``Robotic technologies for
  high-throughput plant phenotyping: Contemporary reviews and future
  perspectives,'' \emph{Frontiers in Plant Science}, vol.~12, 2021.

\bibitem{phenotypedata}
M.~M. Rahaman, D.~Chen, Z.~Gillani, C.~Klukas, and M.~Chen, ``Advanced
  phenotyping and phenotype data analysis for the study of plant growth and
  development,'' \emph{Frontiers in Plant Science}, vol.~6, 2015.

\bibitem{JiangyuReview}
Y.~Jiang and C.~Li, ``Convolutional neural networks for image-based
  high-throughput plant phenotyping: A review,'' \emph{Plant Phenomics}, vol.
  2020, 2020.

\bibitem{PhenotypingBotReview}
R.~Xu and C.~Li, ``A review of high-throughput field phenotyping systems:
  Focusing on ground robots,'' \emph{Plant Phenomics}, vol. 2022, 2022.

\bibitem{genotyping}
S.~Moeinizade, G.~Hu, L.~Wang, and P.~S. Schnable, ``{Optimizing Selection and
  Mating in Genomic Selection with a Look-Ahead Approach: An Operations
  Research Framework},'' \emph{G3 Genes|Genomes|Genetics}, vol.~9, no.~7, pp.
  2123--2133, 07 2019.

\bibitem{genomeselection}
L.~Wang, G.~Zhu, W.~Johnson, and M.~Kher, ``Three new approaches to genomic
  selection,'' \emph{Plant Breeding}, vol. 137, no.~5, pp. 673--681, 2018.

\bibitem{CharlieLiReview}
J.~Iqbal, R.~Xu, H.~Halloran, and C.~Li, ``Development of a multi-purpose
  autonomous differential drive mobile robot for plant phenotyping and soil
  sensing,'' \emph{Electronics}, vol.~9, no.~9, 2020.

\bibitem{agronomy9050258}
A.~Chawade, J.~van Ham, H.~Blomquist, O.~Bagge, E.~Alexandersson, and R.~Ortiz,
  ``High-throughput field-phenotyping tools for plant breeding and precision
  agriculture,'' \emph{Agronomy}, vol.~9, no.~5, 2019.

\bibitem{platforms}
A.~Mazis, S.~D. Choudhury, P.~B. Morgan, V.~Stoerger, J.~Hiller, Y.~Ge, and
  T.~Awada, ``Application of high-throughput plant phenotyping for assessing
  biophysical traits and drought response in two oak species under controlled
  environment,'' \emph{Forest Ecology and Management}, vol. 465, p. 118101,
  2020.

\bibitem{hardwaresoftwarereview}
F.~Solimani, A.~Cardellicchio, M.~Nitti, A.~Lako, G.~Dimauro, and V.~RenÃ²,
  ``A systematic review of effective hardware and software factors affecting
  high-throughput plant phenotyping,'' \emph{Information}, vol.~14, no.~4,
  2023.

\bibitem{ISUwork}
T.~Gao, H.~Emadi, H.~Saha, J.~Zhang, A.~Lofquist, A.~Singh,
  B.~Ganapathysubramanian, S.~Sarkar, A.~K. Singh, and S.~Bhattacharya, ``A
  novel multirobot system for plant phenotyping,'' \emph{Robotics}, vol.~7,
  no.~4, 2018.

\bibitem{Onlinedetection}
M.~Jenkins and G.~Kantor, ``Online detection of occluded plant stalks for
  manipulation,'' in \emph{2017 IEEE/RSJ International Conference on
  Intelligent Robots and Systems (IROS)}, 2017, pp. 5162--5167.

\bibitem{MARS}
R.~Xu and C.~Li, ``A modular agricultural robotic system (mars) for precision
  farming: Concept and implementation,'' \emph{Journal of Field Robotics},
  vol.~39, no.~4, pp. 387--409, 2022.

\bibitem{Sensorreview}
Y.~Qiao, J.~Valente, D.~Su, Z.~Zhang, and D.~He, ``Editorial: Ai, sensors and
  robotics in plant phenotyping and precision agriculture,'' \emph{Frontiers in
  Plant Science}, vol.~13, 2022.

\bibitem{Darwin_2021}
B.~Darwin, P.~Dharmaraj, S.~Prince, D.~E. Popescu, and D.~J. Hemanth,
  ``Recognition of bloom/yield in crop images using deep learning models for
  smart agriculture: A review,'' \emph{Agronomy}, vol.~11, no.~4, p. 646, mar
  2021.

\bibitem{objdetectionreview}
G.~Farjon, H.~Liu, and Y.~Edan, ``Deep-learning-based counting methods,
  datasets, and applications in agriculture: a review,'' \emph{Precision
  Agriculture}, vol.~24, 2023.

\bibitem{RCNN}
R.~Girshick, J.~Donahue, T.~Darrell, and J.~Malik, ``Rich feature hierarchies
  for accurate object detection and semantic segmentation,'' in \emph{2014 IEEE
  Conference on Computer Vision and Pattern Recognition (CVPR)}, 2014, pp.
  580--587.

\bibitem{YOLO}
J.~Redmon, S.~Divvala, R.~Girshick, and A.~Farhadi, ``You only look once:
  Unified, real-time object detection,'' in \emph{2016 IEEE Conference on
  Computer Vision and Pattern Recognition (CVPR)}, 2016, pp. 779--788.

\bibitem{DeepFruit}
I.~Sa, Z.~Ge, F.~Dayoub, B.~Upcroft, T.~Perez, and C.~McCool, ``Deepfruits: A
  fruit detection system using deep neural networks,'' \emph{Sensors}, vol.~16,
  no.~8, 2016.

\bibitem{YOLOforFruit}
K.~Bresilla, G.~D. Perulli, A.~Boini, B.~Morandi, L.~Corelli~Grappadelli, and
  L.~Manfrini, ``Single-shot convolution neural networks for real-time fruit
  detection within the tree,'' \emph{Frontiers in Plant Science}, vol.~10,
  2019.

\bibitem{fewshotlearning}
N.~Ragu and J.~Teo, ``Object detection and classification using few-shot
  learning in smart agriculture: A scoping mini review,'' \emph{Frontiers in
  Sustainable Food Systems}, vol.~6, 2023.

\bibitem{Agimagesynthesis}
N.~Giakoumoglou, E.~Pechlivani, and D.~Tzovaras, ``Generate-paste-blend-detect:
  Synthetic dataset for object detection in the agriculture domain,''
  \emph{Smart Agricultural Technology}, vol.~5, p. 100258, 05 2023.

\bibitem{goodfellow2020generative}
I.~Goodfellow, J.~Pouget-Abadie, M.~Mirza, B.~Xu, D.~Warde-Farley, S.~Ozair,
  A.~Courville, and Y.~Bengio, ``Generative adversarial networks,''
  \emph{Communications of the ACM}, vol.~63, pp. 139--144, 2020.

\bibitem{GANmodel1}
C.~Karam, M.~Awad, Y.~Abou~Jawdah, N.~Ezzeddine, and A.~Fardoun, ``Gan-based
  semi-automated augmentation online tool for agricultural pest detection: A
  case study on whiteflies,'' \emph{Frontiers in Plant Science}, vol.~13, 2022.

\bibitem{GANmodel2}
L.~Bi and G.~Hu, ``Improving image-based plant disease classification with
  generative adversarial network under limited training set,'' \emph{Frontiers
  in Plant Science}, vol.~11, 2020.

\bibitem{kleeberger2020survey}
K.~Kleeberger, R.~Bormann, W.~Kraus, and M.~F. Huber, ``A survey on
  learning-based robotic grasping,'' \emph{Current Robotics Reports}, vol.~1,
  no.~4, pp. 239--249, 2020.

\bibitem{4}
K.~Bousmalis, N.~Silberman, D.~Dohan, D.~Erhan, and D.~Krishnan, ``Unsupervised
  pixel-level domain adaptation with generative adversarial networks,'' in
  \emph{CVPR}, 2017, pp. 3722--3731.

\bibitem{71}
X.~B. Peng, M.~Andrychowicz, W.~Zaremba, and P.~Abbeel, ``Sim-to-real transfer
  of robotic control with dynamics randomization,'' in \emph{2018 ICRA}.\hskip
  1em plus 0.5em minus 0.4em\relax IEEE, 2018, pp. 3803--3810.

\bibitem{domainadaptation}
V.~M. Patel, R.~Gopalan, R.~Li, and R.~Chellappa, ``Visual domain adaptation: A
  survey of recent advances,'' \emph{IEEE signal processing magazine}, vol.~32,
  no.~3, pp. 53--69, 2015.

\bibitem{retinaGAN}
D.~Ho, K.~Rao, Z.~Xu, E.~Jang, M.~Khansari, and Y.~Bai, ``Retinagan: An
  object-aware approach to sim-to-real transfer,'' in \emph{2021 ICRA}.\hskip
  1em plus 0.5em minus 0.4em\relax IEEE, 2021, pp. 10\,920--10\,926.

\bibitem{rao2020rl}
K.~Rao, C.~Harris, A.~Irpan, S.~Levine, J.~Ibarz, and M.~Khansari,
  ``Rl-cyclegan: Reinforcement learning aware simulation-to-real,'' in
  \emph{CVPR}, 2020, pp. 11\,157--11\,166.

\bibitem{DTCycleGAN}
D.~Liu, Y.~Chen, and Z.~Wu, ``Digital twin (dt)-cyclegan: Enabling zero-shot
  sim-to-real transfer of visual grasping models,'' \emph{IEEE Robotics and
  Automation Letters}, vol.~8, no.~5, pp. 2421--2428, 2023.

\bibitem{Mars-Phenobot}
Z.~Li, R.~Xu, C.~Li, and L.~Fu, ``Simulation of an in-field phenotyping robot:
  System design, vision-based navigation and field mapping,'' in \emph{2022
  ASABE Annual International Meeting}.\hskip 1em plus 0.5em minus 0.4em\relax
  American Society of Agricultural and Biological Engineers, 2022, p.~1.

\bibitem{zhu2017unpaired}
J.-Y. Zhu, T.~Park, P.~Isola, and A.~A. Efros, ``Unpaired image-to-image
  translation using cycle-consistent adversarial networks,'' in \emph{ICCV},
  2017, pp. 2223--2232.

\bibitem{YOLOv8f}
J.~Solawetz, ``What is yolov8? the ultimate guide,''
  \emph{https://blog.roboflow.com/whats-new-in-yolov8/}, 2023.

\bibitem{Thorvald}
L.~Grimstad and P.~J. From, ``The thorvald ii agricultural robotic system,''
  \emph{Robotics}, vol.~6, no.~4, 2017.

\bibitem{Scanalyzer}
N.~Virlet, K.~Sabermanesh, P.~Sadeghi-Tehran, and M.~Hawkesford, ``Field
  scanalyzer: An automated robotic field phenotyping platform for detailed crop
  monitoring,'' \emph{Functional Plant Biology}, vol.~44, pp. 143--153, 11
  2016.

\bibitem{liu2021review}
M.~Liu, S.~Fang, H.~Dong, and C.~Xu, ``Review of digital twin about concepts,
  technologies, and industrial applications,'' \emph{Journal of Manufacturing
  Systems}, vol.~58, pp. 346--361, 2021.

\bibitem{DTagriculture}
W.~Purcell and T.~Neubauer, ``Digital twins in agriculture: A state-of-the-art
  review,'' \emph{Smart Agricultural Technology}, vol.~3, p. 100094, 2023.

\bibitem{liu2022digital}
Y.~Liu and et~al., ``A digital twin-based sim-to-real transfer for deep
  reinforcement learning-enabled industrial robot grasping,'' \emph{Robotics
  and Computer-Integrated Manufacturing}, vol.~78, p. 102365, 2022.

\bibitem{Smallsize}
X.~Mai, H.~Zhang, and M.~Q.-H. Meng, ``Faster r-cnn with classifier fusion for
  small fruit detection,'' in \emph{2018 IEEE International Conference on
  Robotics and Automation (ICRA)}, 2018, pp. 7166--7172.

\bibitem{ViTpaper}
A.~Dosovitskiy, L.~Beyer, A.~Kolesnikov, D.~Weissenborn, X.~Zhai,
  T.~Unterthiner, M.~Dehghani, M.~Minderer, G.~Heigold, S.~Gelly, J.~Uszkoreit,
  and N.~Houlsby, ``An image is worth 16x16 words: Transformers for image
  recognition at scale,'' \emph{CoRR}, vol. abs/2010.11929, 2020.

\bibitem{ViTPlant}
S.~{Parez}, N.~{Dilshad}, N.~S. {Alghamdi}, T.~M. {Alanazi}, and J.~W. {Lee},
  ``{Visual Intelligence in Precision Agriculture: Exploring Plant Disease
  Detection via Efficient Vision Transformers},'' \emph{Sensors}, vol.~23,
  no.~15, p. 6949, Aug. 2023.

\bibitem{chen2022maskguided}
Y.~Chen, Z.~Xiao, L.~Zhao, L.~Zhang, H.~Dai, D.~W. Liu, Z.~Wu, C.~Li, T.~Zhang,
  C.~Li, D.~Zhu, T.~Liu, and X.~Jiang, ``Mask-guided vision transformer
  (mg-vit) for few-shot learning,'' \emph{arxiv:
  https://arxiv.org/abs/2205.09995}, 2022.

\bibitem{SANTOS2020105247}
T.~T. Santos, L.~L. {de Souza}, A.~A. {dos Santos}, and S.~Avila, ``Grape
  detection, segmentation, and tracking using deep neural networks and
  three-dimensional association,'' \emph{Computers and Electronics in
  Agriculture}, vol. 170, p. 105247, 2020.

\bibitem{AppleDetector}
P.~K. Sekharamantry, F.~Melgani, and J.~Malacarne, ``Deep learning-based apple
  detection with attention module and improved loss function in yolo,''
  \emph{Remote Sensing}, vol.~15, no.~6, 2023.

\bibitem{Citrus}
Y.~Chen, X.~An, S.~Gao, S.~Li, and H.~Kang, ``A deep learning-based vision
  system combining detection and tracking for fast on-line citrus sorting,''
  \emph{Frontiers in Plant Science}, vol.~12, 2021.

\bibitem{lu2020datasetsurvey}
\BIBentryALTinterwordspacing
Y.~Lu and S.~Young, ``A survey of public datasets for computer vision tasks in
  precision agriculture,'' \emph{Computers and Electronics in Agriculture},
  vol. 178, p. 105760, 2020. [Online]. Available:
  \url{https://www.sciencedirect.com/science/article/pii/S0168169920312709}
\BIBentrySTDinterwordspacing

\bibitem{polvara2023dataset}
R.~Polvara, S.~Molina, I.~Hroob, A.~Papadimitriou, K.~Tsiolis, D.~Giakoumis,
  S.~Likothanassis, D.~Tzovaras, G.~Cielniak, and M.~Hanheide, ``Bacchus
  long-term (blt) data set: Acquisition of the agricultural multimodal blt data
  set with automated robot deployment,'' \emph{Journal of Field Robotics},
  2023.

\bibitem{de2022review}
C.~M. de~Melo, A.~Torralba, L.~Guibas, J.~DiCarlo, R.~Chellappa, and
  J.~Hodgins, ``Next-generation deep learning based on simulators and synthetic
  data,'' \emph{Trends in cognitive sciences}, 2022.

\bibitem{tobin2017domain}
J.~Tobin, R.~Fong, A.~Ray, J.~Schneider, W.~Zaremba, and P.~Abbeel, ``Domain
  randomization for transferring deep neural networks from simulation to the
  real world,'' in \emph{2017 IROS}.\hskip 1em plus 0.5em minus 0.4em\relax
  IEEE, 2017, pp. 23--30.

\bibitem{khalifa2022DL-Data-augumentation-review}
N.~E. Khalifa, M.~Loey, and S.~Mirjalili, ``A comprehensive survey of recent
  trends in deep learning for digital images augmentation,'' \emph{Artificial
  Intelligence Review}, pp. 1--27, 2022.

\bibitem{lu2022GAN-agricultural-review}
Y.~Lu, D.~Chen, E.~Olaniyi, and Y.~Huang, ``Generative adversarial networks
  (gans) for image augmentation in agriculture: A systematic review,''
  \emph{Computers and Electronics in Agriculture}, vol. 200, p. 107208, 2022.

\bibitem{valerio2017DCGAN}
M.~Valerio~Giuffrida, H.~Scharr, and S.~A. Tsaftaris, ``Arigan: Synthetic
  arabidopsis plants using generative adversarial network,'' in
  \emph{Proceedings of the IEEE international conference on computer vision
  workshops}, 2017, pp. 2064--2071.

\bibitem{zhu2018gan-leafcounting}
Y.~Zhu, M.~Aoun, M.~Krijn, J.~Vanschoren, and H.~T. Campus, ``Data augmentation
  using conditional generative adversarial networks for leaf counting in
  arabidopsis plants.'' in \emph{BMVC}, 2018, p. 324.

\bibitem{madsen2019wacgan-seedling}
S.~L. Madsen, M.~Dyrmann, R.~N. J{\o}rgensen, and H.~Karstoft, ``Generating
  artificial images of plant seedlings using generative adversarial networks,''
  \emph{Biosystems Engineering}, vol. 187, pp. 147--159, 2019.

\bibitem{madsen2019semi-GAN-seedlings}
S.~L. Madsen, A.~K. Mortensen, R.~N. J{\o}rgensen, and H.~Karstoft,
  ``Disentangling information in artificial images of plant seedlings using
  semi-supervised gan,'' \emph{Remote Sensing}, vol.~11, no.~22, p. 2671, 2019.

\bibitem{zhu2020_cDCGAN-plantvigor}
F.~Zhu, M.~He, and Z.~Zheng, ``Data augmentation using improved cdcgan for
  plant vigor rating,'' \emph{Computers and Electronics in Agriculture}, vol.
  175, p. 105603, 2020.

\bibitem{drees2021CGAN-brassicagrowth}
L.~Drees, L.~V. Junker-Frohn, J.~Kierdorf, and R.~Roscher, ``Temporal
  prediction and evaluation of brassica growth in the field using conditional
  generative adversarial networks,'' \emph{Computers and Electronics in
  Agriculture}, vol. 190, p. 106415, 2021.

\bibitem{shete2020tasselgan}
S.~Shete, S.~Srinivasan, and T.~A. Gonsalves, ``Tasselgan: An application of
  the generative adversarial model for creating field-based maize tassel
  data,'' \emph{Plant Phenomics}, vol. 2020, 2020.

\bibitem{david2020_wheathead}
E.~David, S.~Madec, P.~Sadeghi-Tehran, H.~Aasen, B.~Zheng, S.~Liu,
  N.~Kirchgessner, G.~Ishikawa, K.~Nagasawa, M.~A. Badhon \emph{et~al.},
  ``Global wheat head detection (gwhd) dataset: a large and diverse dataset of
  high-resolution rgb-labelled images to develop and benchmark wheat head
  detection methods,'' \emph{Plant Phenomics}, 2020.

\bibitem{hartley2021cycleGAN-wheathead}
Z.~K. Hartley and A.~P. French, ``Domain adaptation of synthetic images for
  wheat head detection,'' \emph{Plants}, vol.~10, no.~12, p. 2633, 2021.

\bibitem{rahnemoonfar2017tomatocount}
M.~Rahnemoonfar and C.~Sheppard, ``Deep count: fruit counting based on deep
  simulated learning,'' \emph{Sensors}, vol.~17, no.~4, p. 905, 2017.

\bibitem{dwibedi2017cut}
D.~Dwibedi, I.~Misra, and M.~Hebert, ``Cut, paste and learn: Surprisingly easy
  synthesis for instance detection,'' in \emph{Proceedings of the IEEE
  international conference on computer vision}, 2017, pp. 1301--1310.

\bibitem{picon2022weedsandcorn}
A.~Picon, M.~G. San-Emeterio, A.~Bereciartua-Perez, C.~Klukas, T.~Eggers, and
  R.~Navarra-Mestre, ``Deep learning-based segmentation of multiple species of
  weeds and corn crop using synthetic and real image datasets,''
  \emph{Computers and Electronics in Agriculture}, vol. 194, p. 106719, 2022.

\bibitem{wang2022flowerdetection}
C.~Wang, Y.~Wang, S.~Liu, G.~Lin, P.~He, Z.~Zhang, and Y.~Zhou, ``Study on pear
  flowers detection performance of yolo-pefl model trained with synthetic
  target images,'' \emph{Frontiers in Plant Science}, vol.~13, p. 911473, 2022.

\bibitem{qiu2016gameengine}
W.~Qiu and A.~Yuille, ``Unrealcv: Connecting computer vision to unreal
  engine,'' in \emph{Computer Vision--ECCV 2016 Workshops: Amsterdam, The
  Netherlands, October 8-10 and 15-16, 2016, Proceedings, Part III 14}.\hskip
  1em plus 0.5em minus 0.4em\relax Springer, 2016, pp. 909--916.

\bibitem{andujar2018photogrammetry}
D.~And{\'u}jar, M.~Calle, C.~Fern{\'a}ndez-Quintanilla, {\'A}.~Ribeiro, and
  J.~Dorado, ``Three-dimensional modeling of weed plants using low-cost
  photogrammetry,'' \emph{Sensors}, vol.~18, no.~4, p. 1077, 2018.

\bibitem{vierbergen2023sim2realflower}
W.~Vierbergen, A.~Willekens, D.~Dekeyser, S.~Cool \emph{et~al.}, ``Sim2real
  flower detection towards automated calendula harvesting,'' \emph{Biosystems
  Engineering}, vol. 234, pp. 125--139, 2023.

\bibitem{cieslak2022L-system}
M.~Cieslak, N.~Khan, P.~Ferraro, R.~Soolanayakanahally, S.~J. Robinson,
  I.~Parkin, I.~McQuillan, and P.~Prusinkiewicz, ``L-system models for
  image-based phenomics: case studies of maize and canola,'' \emph{in silico
  Plants}, vol.~4, no.~1, p. diab039, 2022.

\bibitem{Iqbal_2020}
J.~Iqbal, R.~Xu, S.~Sun, and C.~Li, ``Simulation of an autonomous mobile robot
  for {LiDAR}-based in-field phenotyping and navigation,'' \emph{Robotics},
  vol.~9, no.~2, p.~46, jun 2020.

\bibitem{RANSAC}
M.~A. Fischler and R.~C. Bolles, ``Random sample consensus: A paradigm for
  model fitting with applications to image analysis and automated
  cartography,'' \emph{Commun. ACM}, vol.~24, no.~6, p. 381–395, jun 1981.

\bibitem{bommasani2021opportunities}
R.~Bommasani, D.~A. Hudson, E.~Adeli, R.~Altman, S.~Arora, S.~von Arx, M.~S.
  Bernstein, J.~Bohg, A.~Bosselut, E.~Brunskill, E.~Brynjolfsson, S.~Buch,
  D.~Card, R.~Castellon, N.~Chatterji, A.~Chen, K.~Creel, J.~Q. Davis,
  D.~Demszky, C.~Donahue, M.~Doumbouya, E.~Durmus, S.~Ermon, J.~Etchemendy,
  K.~Ethayarajh, L.~Fei-Fei, C.~Finn, T.~Gale, L.~Gillespie, K.~Goel,
  N.~Goodman, S.~Grossman, N.~Guha, T.~Hashimoto, P.~Henderson, J.~Hewitt,
  D.~E. Ho, J.~Hong, K.~Hsu, J.~Huang, T.~Icard, S.~Jain, D.~Jurafsky,
  P.~Kalluri, S.~Karamcheti, G.~Keeling, F.~Khani, O.~Khattab, P.~W. Koh,
  M.~Krass, R.~Krishna, R.~Kuditipudi, A.~Kumar, F.~Ladhak, M.~Lee, T.~Lee,
  J.~Leskovec, I.~Levent, X.~L. Li, X.~Li, T.~Ma, A.~Malik, C.~D. Manning,
  S.~Mirchandani, E.~Mitchell, Z.~Munyikwa, S.~Nair, A.~Narayan, D.~Narayanan,
  B.~Newman, A.~Nie, J.~C. Niebles, H.~Nilforoshan, J.~Nyarko, G.~Ogut, L.~Orr,
  I.~Papadimitriou, J.~S. Park, C.~Piech, E.~Portelance, C.~Potts,
  A.~Raghunathan, R.~Reich, H.~Ren, F.~Rong, Y.~Roohani, C.~Ruiz, J.~Ryan,
  C.~Ré, D.~Sadigh, S.~Sagawa, K.~Santhanam, A.~Shih, K.~Srinivasan,
  A.~Tamkin, R.~Taori, A.~W. Thomas, F.~Tramèr, R.~E. Wang, W.~Wang, B.~Wu,
  J.~Wu, Y.~Wu, S.~M. Xie, M.~Yasunaga, J.~You, M.~Zaharia, M.~Zhang, T.~Zhang,
  X.~Zhang, Y.~Zhang, L.~Zheng, K.~Zhou, and P.~Liang, ``On the opportunities
  and risks of foundation models,'' 2021.

\bibitem{zhou2023comprehensive}
C.~Zhou, Q.~Li, C.~Li, J.~Yu, Y.~Liu, G.~Wang, K.~Zhang, C.~Ji, Q.~Yan, L.~He
  \emph{et~al.}, ``A comprehensive survey on pretrained foundation models: A
  history from bert to chatgpt,'' \emph{https://arxiv.org/abs/2108.07258},
  2023.

\bibitem{kirillov2023segment}
A.~Kirillov, E.~Mintun, N.~Ravi, H.~Mao, C.~Rolland, L.~Gustafson, T.~Xiao,
  S.~Whitehead, A.~C. Berg, W.-Y. Lo \emph{et~al.}, ``Segment anything,''
  \emph{arXiv preprint arXiv:2304.02643}, 2023.

\end{thebibliography}
\addtolength{\textheight}{-12cm} 
\end{document}